\documentclass[10pt,twocolumn,letterpaper]{article}

\usepackage[pagenumbers]{cvpr} 
\usepackage{times}  
\usepackage{helvet}  
\usepackage{courier}  
\usepackage[hyphens]{url}  
\usepackage{graphicx} 
\usepackage{amsmath}
\usepackage{amsfonts}
\usepackage{booktabs}
\usepackage{xcolor}
\definecolor{cvprblue}{rgb}{0.21,0.49,0.74}
\usepackage[pagebackref,breaklinks,colorlinks,allcolors=cvprblue]{hyperref}
\usepackage{newfloat}
\usepackage{algorithm}
\usepackage{algorithmic}

\def\paperID{*****} 
\def\confName{CVPR}
\def\confYear{2026}

\title{Vectorized Video Representation with Easy Editing via Hierarchical Spatio-Temporally Consistent Proxy Embedding}

\author{
Ye Chen$^{1\ast}$~~~~Liming Tan$^{1\ast}$~~~~Yupeng Zhu$^{1}$~~~~Yuanbin Wang$^{1}$~~~~Bingbing Ni$^{1,2\dag}$\\
$^{1}$Shanghai Jiao Tong University, Shanghai 200240, China\\$^{2}$USC-SJTU Institute of Cultural and Creative Industry\\
\tt\small \{chenye123, spinningfever, nibingbing\}@sjtu.edu.cn
}

\begin{document}
\maketitle
\renewcommand{\thefootnote}{}
\footnotetext{$^{\ast}$Equal contributions}
\footnotetext{$^{\dag}$Corresponding author:Bingbing Ni}

\begin{abstract}
Current video representations heavily rely on unstable and over-grained priors for motion and appearance modelling, \emph{i.e.}, pixel-level matching and tracking. A tracking error of just a few pixels would lead to the collapse of the visual object representation, not to mention occlusions and large motion frequently occurring in videos. To overcome the above mentioned vulnerability, this work proposes spatio-temporally consistent proxy nodes to represent dynamically changing objects/scenes in the video. On the one hand, the hierarchical proxy nodes have the ability to stably express the multi-scale structure of visual objects, so they are not affected by accumulated tracking error, long-term motion, occlusion, and viewpoint variation. On the other hand, the dynamic representation update mechanism of the proxy nodes adequately leverages spatio-temporal priors of the video to mitigate the impact of inaccurate trackers, thereby effectively handling drastic changes in scenes and objects. Additionally, the decoupled encoding manner of the shape and texture representations across different visual objects in the video facilitates controllable and fine-grained appearance editing capability. Extensive experiments demonstrate that the proposed representation achieves high video reconstruction accuracy with fewer parameters and supports complex video processing tasks, including video in-painting and keyframe-based temporally consistent video editing.
\end{abstract}    
\section{Introduction}

Interactive video editing are critical in multimedia industry, including advertising, film-making, and virtual reality, \emph{etc.}, enabling enriched content creation and immersive experiences~\cite{irani1995mosaic,irani2002video,bovik2010handbook,bonneel2014interactive,wu2023tune}. 
Recent AIGC-based video editing approaches~\cite{wu2023tune, chai2023stablevideo, zhang2024moonshot, yang2025videograin} attempt to map multi-modal codes in the latent space to the pixel domain for manipulation.
Lacking explicit/direct alignment between the latent space and semantic objects in the video pixel space, these approaches are NOT controllable or stable with respect to users' prompt, while sampling in the high dimensional space yields large computationally burden~\cite{rombach2022high, mou2024revideo, ma2025magicstick}. To this end, it lies in the heart to construct an advanced video representation, which not only directly bridges user editing instructions to pixel-level modifications, but also enables the stable preservation of fine-grained structural and textural details. This is a critical step toward controllable, high-fidelity video editing.

\begin{figure}[t]
\includegraphics[width=1.0\linewidth]{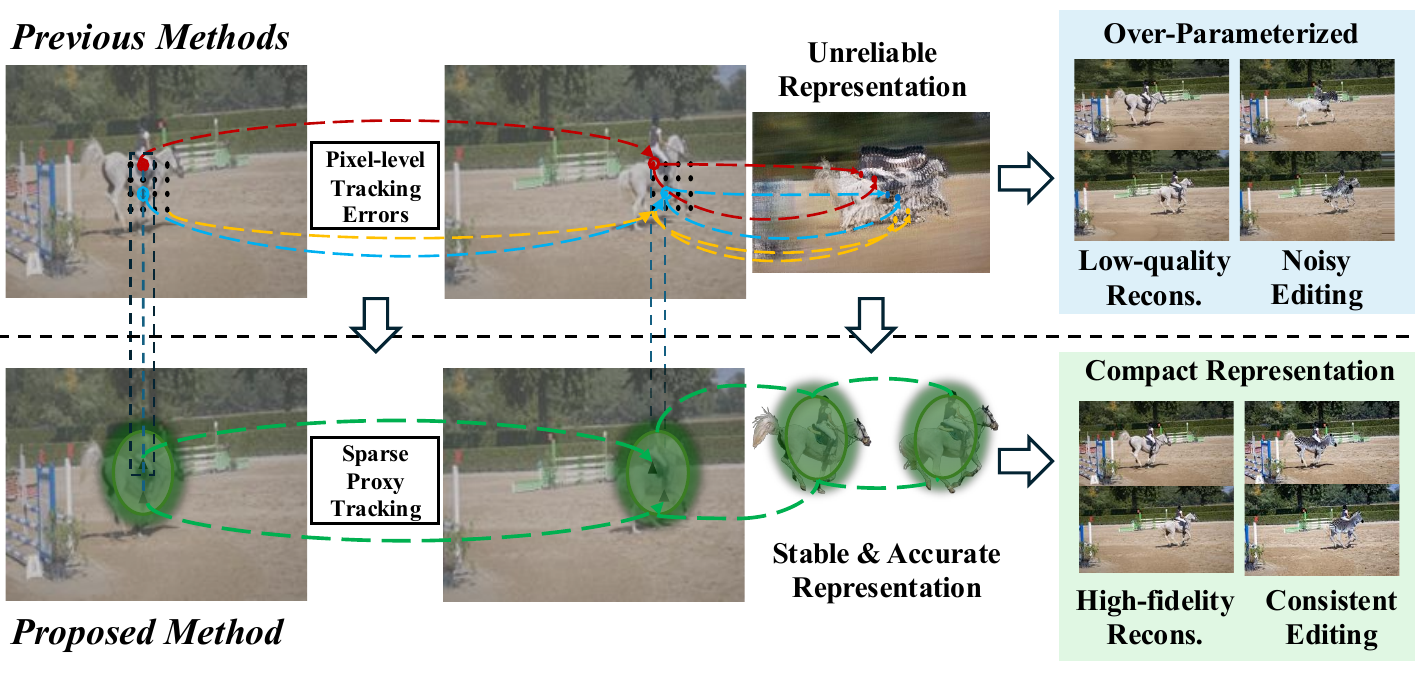} 
\caption{\textbf{Illustration of our Motivation.} Current video representations depend heavily on low-level, unstable priors for motion and appearance modeling, such as pixel-wise matching and tracking, which introduces representation errors and impairs performance of downstream tasks like video editing, especially on videos with occlusions and large motions. This work proposes encoding multi-scale local structures using hierarchical spatio-temporally consistent proxy nodes, and mitigating tracking errors by operating on the trajectories of sparse proxies, which achieves stable and accurate representation and supports complex video processing tasks.}
\label{fig1:motivation}
\end{figure}

Current video representations could be broadly categorized into two paradigms. The first focuses on 2D/2.5D-level representations~\cite{rav2008unwrap, kasten2021layered, bar2022text2live, ye2022deformable, ouyang2024codef}, where pixel-level tracking~\cite{doersch2023tapir, karaev2024cotracker, neoral2024mft} or optical flow estimation~\cite{jabri2020space, teed2020raft,  huang2022flowformer} is used to aggregate temporally aligned pixels into a unified canonical structure (\emph{e.g.}, atlases~\cite{kasten2021layered} or canonical images~\cite{ouyang2024codef}). Edits made to this structure are then propagated to the entire video using estimated optical flow. While these approaches offer explicit temporal consistency, their performance is heavily constrained by the accuracy of the tracker, struggling with occlusion, large-scale motion, and nonrigid deformation, making them unsuitable for complex in-the-wild videos. In addition, due to the inherent distortions in the estimated atlases or canonical images, such methods suffer from compromised semantic integrity, which limits their effectiveness when applying image processing techniques that assume a natural and coherent visual domain for video processing~\cite{ouyang2024codef}.

The second line of work exploits the underlying 3D priors in videos, leveraging monocular depth estimation or 3D reconstruction techniques to explicitly recover the scene geometry and perform editing via manipulations in 3D space~\cite{chen2011video, fruhstuck2023vive3d, sun2024splatter, gu2025diffusion}. For example, the recent work VGR~\cite{sun2024splatter} models video appearance using 3D Gaussians~\cite{3DGS} and imposes monocular priors from 2D foundation models~\cite{yang2024depth, yang2024depthv2} to assign temporally coherent trajectories to these primitives. Thanks to the expressive nature of Gaussians, VGR can reconstruct high-quality video scenes. However, these methods are better suited for domains with accurate 3D priors (\emph{e.g.}, camera poses or object trajectories), such as game production, since recovering precise and temporally consistent 3D information from in-the-wild monocular videos is a highly ill-posed problem.
As a result, VGR performs poorly on videos with large object or camera motion, especially long sequences. Furthermore, due to the limitations of monocular depth estimation, VGR cannot accurately model occlusion relationships and fails to reconstruct video scenes with high fidelity, which significantly limits its applicability to editing tasks such as precise video in-painting.

To overcome these limitations, we introduce a novel video representation framework inspired by parameterized (\emph{i.e.}, vectorization) proxy representations from 2D images~\cite{chen2021learning, muller2022instant, chen2025easy}, where per-frame objects are decomposed into sparse spatial proxy nodes; each node implicitly encodes the shape and texture of its local compact region, enabling stable spatio-temporal propagation preserving fine structures for high-fidelity reconstruction and precise editing due to their decoupled spatial-attribute nature. Specifically, we first decompose the video scene into semantic layers and initialize for each layer a set of proxy nodes, including contour control points and internal geometric points~\cite{hu2019triwild}, to embed local appearance and structure. 
Extending beyond 2D, the core challenge for video lies in establishing a temporally consistent/coherent proxy representation, requiring robust temporal linking of proxy nodes and their embedded visual codes across frames, which is a non-trivial task due to the inherent instability of tracking and optical flow algorithms. 
To mitigate noisy motion estimation, we employ the following strategies. First, instead of pixel-wise cross-frame matching, we propagate proxy nodes through the video. Due to their sparse spatial distribution, proxy nodes are more tolerant of tracking errors than dense pixel matching. Second, we introduce a dynamic proxy node augmentation and propagation mechanism, which adaptively inserts new proxy nodes during forward proxy propagation to compensate for accumulated error and preserve representation integrity. In addition, bi-directional propagation of supplemented nodes allows us to capture multi-scale temporal priors and effectively encode occluded regions. For instance, background regions occluded by a horse in the first frame can be recovered by propagating proxy nodes from the last frame where these regions are visible, enabling precise occlusion reasoning and background completion (refer to Fig.\ref{fig2:overview}). Appearance feature codes are therefore moved along with the corresponding proxy nodes across frames, providing a stable support domain for implicit neural image reconstructions (\emph{i.e.}, implicit mapping function), as the proxy nodes inherently integrate multi-scale local geometry and appearance. The above design ensures the stability and consistency of appearance under large motions and facilitates controllable editing.

Leveraging our efficient distributed proxy-based representation, videos are optimized with only a few minutes. Furthermore, it enables unprecedented high-precision video editing and processing, significantly outperforming prior methods in tasks including: 1) controllable and accurate video in-painting; 2) keyframe-based consistent video editing; and 3) spatio-temporal frame interpolation. Extensive experiments across reconstruction and diverse editing tasks validate its effectiveness and efficiency.
\section{Related Works}
\noindent \textbf{Image Vectorization \& Editing.} Image vectorization aims to utilize parametric primitives to represent images, enabling user-friendly interactive image editing.
With the advent of differentiable rasterization framework~\cite{li2020differentiable}, neural network-based image vectorization approaches~\cite{reddy2021im2vec, ma2022towards, chen2023editable, du2023image, hu2024supersvg} gain significant interests in recent years. LIVE~\cite{ma2022towards} represents a pioneering effort to vectorize images into layer-wise primitives through a dedicated path initialization strategy. Du \emph{et al.}~\cite{du2023image} propose using linear gradients to decompose images into vectorized layers, enabling structured and intuitive image editing. However, these approaches are restricted to simple artistic images and struggle to generalize to complex natural scenes, primarily due to the insufficient texture representation capacity of geometric primitives. In particular, methods~\cite{chen2024towards, chen2025easy} that combine geometric primitives with implicit texture representations extend editable image vectorization to natural images, enhancing both representation capacity and texture stability during editing.
Although these methods achieve remarkable results in image representation and editing, extending them to video representation and consistent editing remains challenging due to the increased complexity of video content.
This paper is inspired by image vectorization methods and aims to achieve efficient video representation and consistent editing.

\noindent \textbf{Video Representation \& Editing.} Early methods~\cite{irani2002video, correa2010dynamic, barnes2010video} are largely based on video mosaics, which attempt to construct a global panorama or reference frame by stitching together multiple frames.
These mosaics serve as a proxy for the entire video, allowing edits made on the mosaic to be propagated across frames.
LNA~\cite{kasten2021layered} extends mosaic-based approaches to complex in-the-wild videos with a layer-wise strategy by jointly optimizing layer-wise atlases and coordinate-to-RGBA mappings constrained by optical flow estimation. Due to the unnatural appearance of the estimated atlases, LNA is unable to leverage advanced image editing techniques to support diverse video editing tasks. CoDeF~\cite{ouyang2024codef} models frame-wise deformations with respect to a canonical field by learning a multi-resolution hash grid, which successfully lifts image algorithms to video editing. Another line of work~\cite{liu2023robust, wu20244d, sun2024splatter, smolak2024vegas} achieves consistent video editing by estimating 3D information from the video. VGR~\cite{sun2024splatter} and VeGaS~\cite{smolak2024vegas} are remarkable works, which propose to represent videos using 3D Gaussians embedded with 3D trajectories with the help of priors from 2D foundation models, enabling effective modeling of occlusions in videos.
However, all above methods are significantly limited by unreliable tracking and 3D estimation, especially under large-scale motions.
Recent years have also seen rapid progress in generative model-based video editing~\cite{wu2023tune, qi2023fatezero, mou2024revideo, jiang2025vace, hu2025hunyuancustom}; yet, pixel-domain probabilistic approaches still struggle to deliver controllable and temporally consistent edits. Our work proposes an efficient video representation that enables consistent editing with reduced reliance on precise tracking.

\section{Methodology}
\begin{figure*}[t]
\centering
\includegraphics[width=1.0\linewidth]{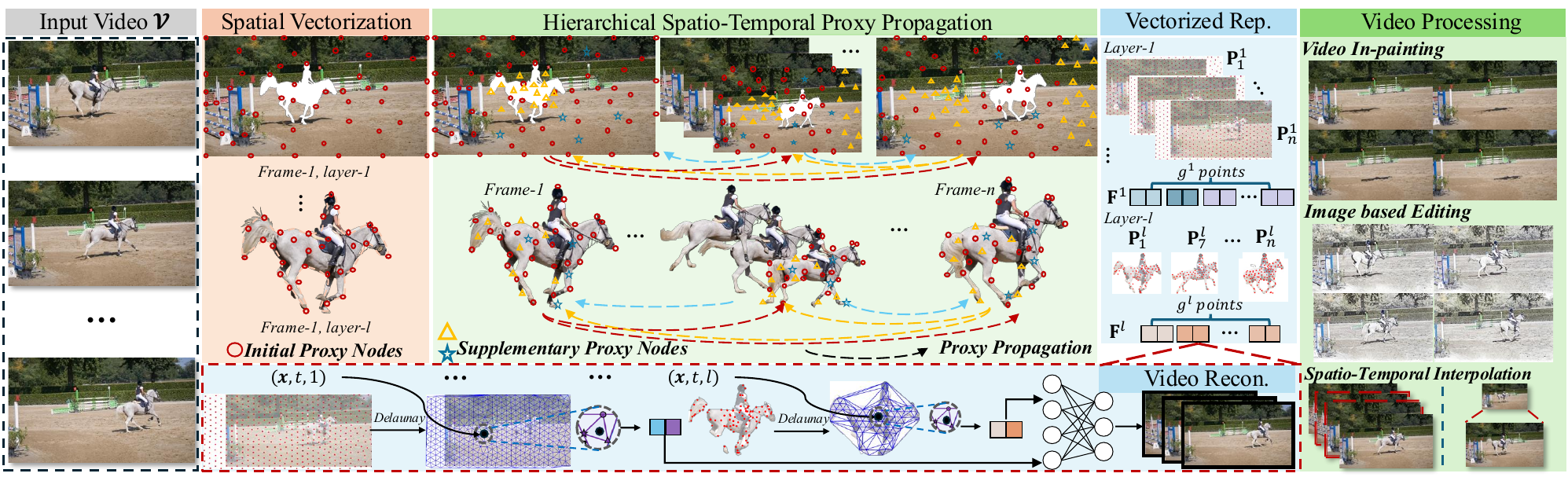} 
\caption{\textbf{Overview of our framework.} We decompose the input video into semantic layers, and then embed video motion and appearance into hierarchical spatio-temporally consistent proxy nodes through a dynamic proxy node supplementation and propagation mechanism. Our representation enables various video processing tasks. The dashed arrows indicate the propagation of corresponding color-coded supplementary nodes from current frame to target frame via the tracking algorithm.}
\label{fig2:overview}
\vspace{-0.4cm}
\end{figure*}

\subsection{Overview}
The overview framework is illustrated in Fig.~\ref{fig2:overview}. We parameterize any input video $\small \mathcal{V}=\{\mathbf{I}_1,\mathbf{I}_2,...,\mathbf{I}_n\}$ into implicit embeddings distributed at multi-layer spatio-temporal proxy nodes and a coord-to-RGB decoding function $\phi_\theta$:
\begin{equation}
\small
    \mathcal{V} \sim \{\{\mathbf{G}^1, \mathbf{G}^2, ..., \mathbf{G}^l\}, \theta\},
\end{equation}
where $l$ denotes the number of semantic layers and $\small \mathbf{G}^i=[\mathbf{P}^i, \mathbf{F}^i]$ represents proxy nodes of each layer, $\small \mathbf{P}^i \in \mathbb{R}^{g^{i}\times 2n}$ denotes the positions of the nodes across all frames, and $\small \mathbf{F}^i \in \mathbb{R}^{g^{i}\times c}$ represents the texture codes distributed at proxy nodes with $n$ denoting the number of frames, $g^{i}$ representing the node number of each layer and $c$ denoting the dimension of texture codes. Details are elaborated next.

\subsection{Video Spatial Vectorization}
To spatially disentangle the video for finer representations and more intuitive editing, we propose to perform video spatial vectorization. Specifically, we initialize the spatial structure of the video using Grounded SAM2~\cite{ravi2024sam, liu2024grounding}, which decompose the video into a set of masks:
\begin{equation}
\small
    \mathcal{V} \rightarrow [\mathbf{M}^1, \mathbf{M}^2,..., \mathbf{M}^l], 
\end{equation}
where 
\begin{equation}
\small
    \mathbf{M}^i = [\mathbf{M}_{t_s^{i}}^{i}, \mathbf{M}_{t_s^{i}+1}^{i},..., \mathbf{M}_{t_e^{i}}^{i}],
\end{equation}
where $t_s^{i}$ and $t_e^{i}$ denote the frame numbers when the $i$-th semantic layer appears and disappears, respectively. Note that Grounded SAM2 may exhibit temporal instability on video object tracking. However, this does not affect our algorithm as we only need to identify the initial frame corresponding to the object of editing interest, while the terminal frame can generally be set as the last frame of the video.

For each decomposed layer, we select the frame where it first appears (represented as $\mathbf{M}^i_{t^{i}_{s}}$) and employ VTracer~\cite{VisionCortexVTracer2023} to fit its edges and derive a series of edge control points that capture its structural information:
\begin{equation}
\small
    \mathbf{P}^{i,edge}_{t^{i}_{s}} = \textit{\small Vtracer}(\mathbf{M}^i_{t^{i}_{s}}).
\end{equation}
Inspired by image vectorization algorithms~\cite{chen2025easy, hu2019triwild}, to further extract the fine geometric structure within each layer, we use the Sobel operator to calculate the gradient of each pixel within the layer. We then sample a series of internal control points in descending order of gradient, which, along with the edge control points, specify the spatial positions of the proxy nodes for the corresponding semantic layer when it first appears:
\begin{equation}
\small
    \mathbf{P}^{i,0}_{t^{i}_{s}} = \mathbf{P}^{i,edge}_{t^{i}_{s}} \cup \textit{\small Sobel\_Sample}(\mathbf{I}_{t^{i}_{s}}\cdot\mathbf{M}^i_{t^{i}_{s}}).
\end{equation}

The video space is therefore decomposed into multi-layer spatio-temporal control points (\emph{i.e.}, $\small \cup_{i=1}^{l} \mathbf{P}^{i,0}_{t^{i}_{s}}$). These points function as visual feature embedding anchors (\emph{i.e.}, named as proxy nodes) for the geometric structure, motion dynamics, and visual appearance of various video objects. 

\subsection{Hierarchical Spatio-temporal Proxy Propagation}
After acquiring the spatio-temporal positions of the initial proxy nodes for each semantic layer, the primary goal is to comprehensively encode all spatio-temporal attributes (\emph{i.e.}, motion and temporally varying appearance) into these proxy nodes such that realistic and globally consistent video reconstructing/editing can be achieved by solely decoding/modifying the attribute parameters distributed on the proxy nodes. Most existing methods~\cite{kasten2021layered, ouyang2024codef, sun2024splatter} rely on off-the-shelf trackers~\cite{teed2020raft, karaev2024cotracker} to estimate dense pixel trajectories for temporal aggregation. However, they often neglect multi-scale temporal structures and are highly dependent on tracker accuracy, making them unstable in cases of large motions and frequent occlusions.

Instead, this work proposes to \emph{coarsely} track proxy nodes. The key philosophy is: unlike pixel-level features, our proxy nodes robustly encode local structure and appearance, remaining largely unaffected by sub-pixel tracking errors. Specifically, for the $i$-th layer, we firstly employ CoTraker~\cite{karaev2024cotracker} to propagate $\small \mathbf{P}^{i,0}_{t^{i}_{s}}$ from frame $\small t^{i}_{s}$ to $\small t^{i}_{e}$ and obtain the initial temporal trajectories:
\begin{equation}
\small
    \mathbf{P}^{i,0} = [\mathbf{P}^{i,0}_{t^{i}_{s}}, \mathbf{P}^{i,0}_{t^{i}_{s}+1},..., \mathbf{P}^{i,0}_{t^{i}_{e}}].
\end{equation}
Moreover, to deal with tracking error accumulation over long sequences as well as large appearance change, we also propose a dynamic proxy node augmentation and propagation module to reinforce multi-scale temporal priors from the video, leading to a more accurate re-distribution of proxy nodes. More concretely, after the first round of propagation, starting from the last frame, we compute for each pixel in layer $i$ of this frame (\emph{i.e.}, $\small \mathbf{I}_{t^{i}_{e}}\cdot\mathbf{M}^i_{t^{i}_{e}}$), the distance $d$ to its nearest proxy node. We define each pixel with $d\geq\epsilon_d$ as a non-proxy point. We then iteratively sample new nodes from the non-proxy points and update all $d$ values until no non-proxy points remain.
Then, we perform a reverse propagation of these supplementary points starting from the current frame (\emph{i.e.}, the last frame) back to the first frame. We sequentially perform the above proxy node supplementation and bidirectional propagation starting from the second frame, in order to fully capture the temporal hierarchies of the video. Finally, the trajectories of all proxy nodes of layer $i$ can be represented as:
\begin{equation}
\small
    \mathbf{P}^{i} = [\mathbf{P}^{i}_{t^{i}_{s}}, \mathbf{P}^{i}_{t^{i}_{s}+1},..., \mathbf{P}^{i}_{t^{i}_{e}}],
\end{equation}
with
\begin{equation}
\small
    \mathbf{P}^{i}_{t\ast} = \cup_{j=0}^{k}\mathbf{P}^{i,j}_{t\ast},
\end{equation}
where $k$ represents the number of supplementation rounds and $\small \mathbf{P}^{i,j}_{t\ast}$ represents the positions of the nodes supplemented in the $j$-th round as propagated to frame $t\ast$. 

Note that due to the existence of occlusions, the proxy nodes supplemented in a particular frame may not always have semantically corresponding points in every frame during propagation, which may introduce unacceptable noise into the motion information encoded by proxy nodes. Take background nodes as an example: if a background region visible in frame $t_e$ is occluded by a horse in frame $t_s$, then the nodes supplemented in frame $t_e$ will not have semantically consistent counterparts in frame $t_s$. However, we observe that, due to the continuity modelled by neural networks, the trajectories of such points can be approximated by a weighted average of their neighboring points that are present in both frame $t_e$ and $t_s$. Consequently, propagating these points from frame $t_e$ to frame $t_s$ remains meaningful, as they naturally align with the background regions occluded by the horse in frame $t_s$.
Our method efficiently reduces tracking errors' impact on video representations and uses temporal priors to fill occluded regions, enabling more flexible and powerful editing tasks like video in-painting.

\subsection{Vectorized Video Representation Optimization}
After completing dynamic proxy node generation and propagation, we obtain the trajectories of all proxy nodes for each semantic layer, which encode the overall video motion structure (in a sparse yet robust manner). Next, we embed video appearance into the proxy nodes in an implicit manner following distributed implicit representation methods~\cite{muller2022instant, chen2024towards} in the image domain.

Specifically, for each layer $i$ with $g^{i}$ proxy nodes (\emph{i.e.}, $\small \mathbf{P}^i \in \mathbb{R}^{g^{i}\times 2n}$), we first distribute randomly initialized texture codes ($\small \mathbf{F}^i \in \mathbb{R}^{g^{i}\times c}$) at all proxy nodes. It is noted the attached feature remains unchanged even though the position of the corresponding node varies across frames. We render $\small \mathbf{F}^i$ onto each pixel of every frame by following the trajectories of the proxy nodes, and optimize $\small \mathbf{F}^i$ using an $L_2$ loss with respect to the original video. To ensure a stable mapping between proxy nodes and pixel values, we employ a per-frame triangulation strategy, assigning each proxy node the responsibility of reconstructing pixels within its associated triangle. In more detail, we first perform Delaunay triangulation on all proxy nodes each layer $i$ of every frame to obtain a set of triangles:
\begin{align}
\small
    \mathbf{T}^{i}_{t\ast} &= \textit{Delaunay}(\mathbf{P}^{i}_{t\ast}), \\
    \mathbf{T}^{i} =& [\mathbf{T}^{i}_{t^{i}_{s}}, \mathbf{T}^{i}_{t^{i}_{s}+1},..., \mathbf{T}^{i}_{t^{i}_{e}}].
\end{align}
Note that triangulations across consecutive frames are generally computed independently because we focus solely on the temporal consistency of the proxy nodes (\emph{i.e.}, motion), without enforcing topological constraints within each semantic layer. This flexibility allows us to better model objects undergoing topology or shape changes. For a given pixel point $\boldsymbol{x}$ in layer $i$ of frame $t$, we then identify the triangle it lies in using barycentric coordinates, which are subsequently used as interpolation weights to compute the pixel's corresponding feature. The above process can be denoted as:
\begin{equation}
\small
    f_{t,\boldsymbol{x}}^{i} = \sum_{k=1}^{3}\lambda^{i,k}_{t,\boldsymbol{x}}\cdot \mathbf{F}^{i,k},
\end{equation}
where $\small \mathbf{F}^{i,k}$ and $\small \lambda^{i,k}_{t,\boldsymbol{x}}$ denote the texture codes of corresponding vertices of $\small \mathbf{T}^{i}_{t}$ and associated barycentric weights of $\boldsymbol{x}$, as identified via the Barycentric Coordinate Test. Then the texture value at point $\boldsymbol{x}$ is decoded to RGB value with a decoding function $\phi_\theta$. To efficiently capture the spatio-temporal variations in appearance (such as shadows) observed in video sequences, we also incorporate the spatio-temporal coordinate ($t, \boldsymbol{x}$) as an additional input to the function, which can be described as:
\begin{equation}
\small
    \hat{I}_{t,\boldsymbol{x}}^{i} = \phi_\theta(\mathcal{U}_{freq}([f_{t,\boldsymbol{x}}^{i}, t, \boldsymbol{x}])),
    \label{eq:recons_pixel}
\end{equation}
where $\small \mathcal{U}_{freq}$ is a encoding function to map feature codes and coordinates into high-frequency space as defined in~\cite{mildenhall2021nerf}. In each iteration of the optimization process, a set of spatio-temporal pixel coordinates $\mathcal{C}$ is randomly sampled from the entire video. The pixel-wise mean squared error is then computed to simultaneously optimize the texture encoding and decoding functions:
\begin{equation}
\small
    \min_{\{\mathbf{F}^1, \mathbf{F}^2,..., \mathbf{F}^l, \theta\}} \sum_{(\boldsymbol{x},t,i)\in\mathcal{C}}||I_{t,\boldsymbol{x}}^{i}-\hat{I}_{t,\boldsymbol{x}}^{i}||_{2}^{2}.
\end{equation}
After optimization, the reconstructed video can be rendered by applying Eqn.~(\ref{eq:recons_pixel}) in parallel to all pixels across all layers and frames. Furthermore, realistic video editing can be easily achieved by adjusting the positions or feature embeddings of proxy nodes in a decoupled, layer-wise manner.

\section{Experiments}
\subsection{Experimental Setups}
\noindent\textbf{Dataset\&Evaluation.}
Experiments are conducted on commonly used benchmark DAVIS~\cite{pont20172017} as well as some videos used by prior works for fair comparisons. 
We evaluate our method on video representation task and video processing tasks (including a.\textbf{video in-painting}; b.\textbf{image-based consistent editing}; c.\textbf{spatio-temporal video interpolation}) and make comparisons with SOTAs of both video representation-based methods(\emph{i.e.}, LNA~\cite{kasten2021layered}, CoDeF~\cite{ouyang2024codef}, VGR~\cite{sun2024splatter}, VeGaS~\cite{smolak2024vegas} and advanced generative model-based methods Inpaint-Anything~\cite{yu2023inpaint} and Vid-Dir~\cite{wang2025videodirector}.

\noindent\textbf{Implementation Details.} We set the threshold for non-proxy point $\epsilon_d$ as $30/\min[h,w]$, where $[h,w]$ is video resolution. The dimension of texture codes $c$ is set to $128$ to balance representation accuracy and efficiency. $\phi_\theta$ is a $8-$layer MLP with hidden dimension $256$ and output
dimension $3$. The frequency number of $\small \mathcal{U}_{freq}$ is set to 9. Each video is optimized for $10000$ steps with Adam optimizer with learning rate $l_r=1e-3$. More details in supplementary materials. 

\subsection{Video Representation}
\begin{figure*}[t]
\centering
\includegraphics[width=0.95\linewidth]{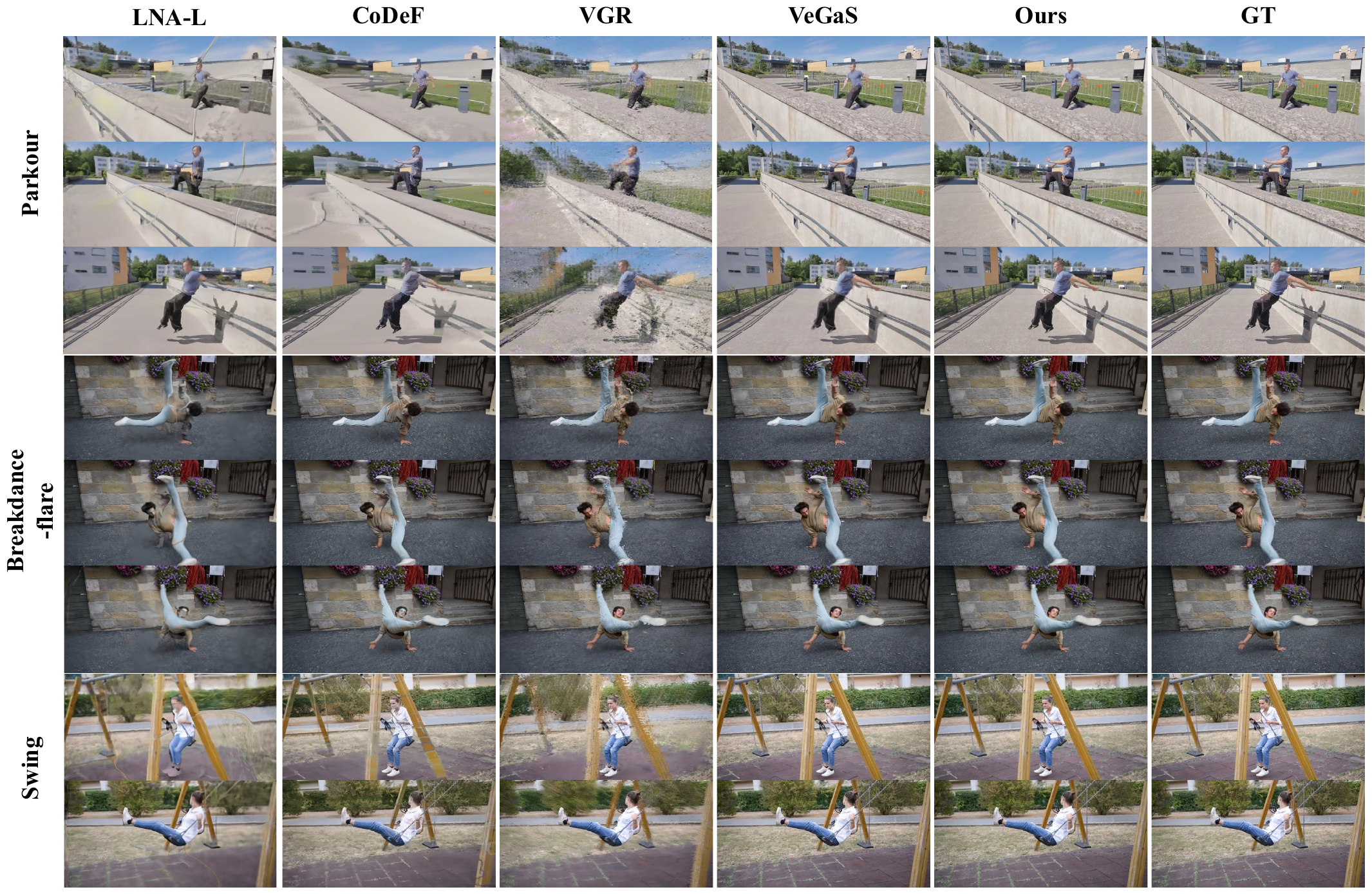} 
\caption{Qualitative comparisons on DAVIS. Our method achieves high-fidelity reconstruction with intricate texture details, especially on complex scenes. Please zoom in for more details. More visualizations are provided in the supplementary material.}
\label{fig3:recons}
\vspace{-0.3cm}
\end{figure*}

On video representation task, we compare our method with advanced video representations. Quantitative results on DAVIS are shown in Tab.\ref{tab:video_recons}. The results of LNA~\cite{kasten2021layered} are computed on a subset of sampled videos due to the extremely long optimization time (please refer to supplementary materials for the sampled list), whereas the metrics for all other methods are averaged over the entire DAVIS video dataset. Note that both LNA and VGR fail to reproduce the results reported in their papers on the whole DAVIS dataset, especially for complex long sequences, which is a widely mentioned issue in their official open-source repositories. Given that the original version of LNA has very few parameters, we increase its capacity (LNA-L) for complex video tasks, but the performance improvement is minimal. We can observe that only VeGaS~\cite{smolak2024vegas} and our method achieve satisfactory reconstruction performance across the full DAVIS dataset, with average PSNR exceeding $30$ dB. However, while VeGaS performs well on the $480$p-version, its performance degrades significantly as the resolution increases. In contrast, our method consistently achieves high-quality video reconstruction even at higher resolutions, thanks to the efficient representation of local structural and textural information via proxy nodes and the adaptive proxy update mechanism. In addition, our approach involves significantly fewer parameters and requires less optimization time, further demonstrating the efficiency of the proxy-node-based representation.

We visualize several challenging video sequences that exhibit large-scale motion, occlusions, and scene changes, with the qualitative results presented in Fig.~\ref{fig3:recons}. LNA, CoDeF, and VGR exhibit clear appearance errors due to their heavy reliance on pixel-level optical flow, which is susceptible to large prediction inaccuracies. While VeGaS delivers acceptable results, it still suffers from aliasing artifacts caused by its use of discrete point-based representations. In contrast, our method minimizes dependence on optical flow by leveraging proxy node representations and a dynamic update mechanism. Additionally, the implicit texture representations embedded in proxy nodes enable stable and accurate appearance modeling, allowing our approach to perform robustly even in complex scenarios.

\begin{table}[t]
    \centering
    \small
    \begin{center}
    \begin{tabular}{l|c|c|c|c|c}
    \toprule
    Method &PSNR$\uparrow$&LPIPS$\downarrow$&SSIM$\uparrow$&Params.$\downarrow$&Time$\downarrow$ \\
    
    \midrule
    \multicolumn{6}{c}{Resolution: $480\times854$} \\
    \midrule
    
    LNA-S&24.43&0.3293&0.6932&\textbf{1.32M}&10h \\
    LNA-L&25.12&0.3087&0.7014&10.0M& \textgreater20h\\
    CoDeF&26.38&0.2274&0.7985&37.9M&30m \\
    VGR&23.97&0.3668&0.6902&300M&1h \\
    VeGaS&32.12&0.1270&\textbf{0.9021}&34.5M&1h \\
    Ours &\textbf{32.58}&\textbf{0.1196}&0.8982&3.17M&\textbf{20m} \\
    
    \midrule
    \multicolumn{6}{c}{Resolution: $1080\times1920$} \\
    \midrule
    
    LNA-S&24.98&0.3162&0.6901&\textbf{1.32M}&10h \\
    LNA-L&25.61&0.2818&0.7103&10.0M& \textgreater20h\\
    
    CoDeF&27.43&0.2218&0.7828&37.9M&\textbf{40m} \\
    VGR&23.42&0.4206&0.6840&350M&2h \\
    VeGaS&31.14&0.1597&0.8876&39.8M&2h \\
    Ours &\textbf{33.49}&\textbf{0.1089}&\textbf{0.9153}&3.21M&\textbf{40m} \\
    
    \bottomrule
    \end{tabular}
    \end{center}
    \vspace{-0.4cm}
    \caption{\textbf{Quantitative results on whole DAVIS.} Our method achieves the best reconstruction results at two different resolutions with very few parameters. The Time is tested on an NVIDIA GeForce RTX 3090 GPU.}
    \label{tab:video_recons}
    \vspace{-0.2cm}
\end{table}

\subsection{Video Processing}
\begin{figure*}[t]
\centering
\includegraphics[width=1.0\linewidth]{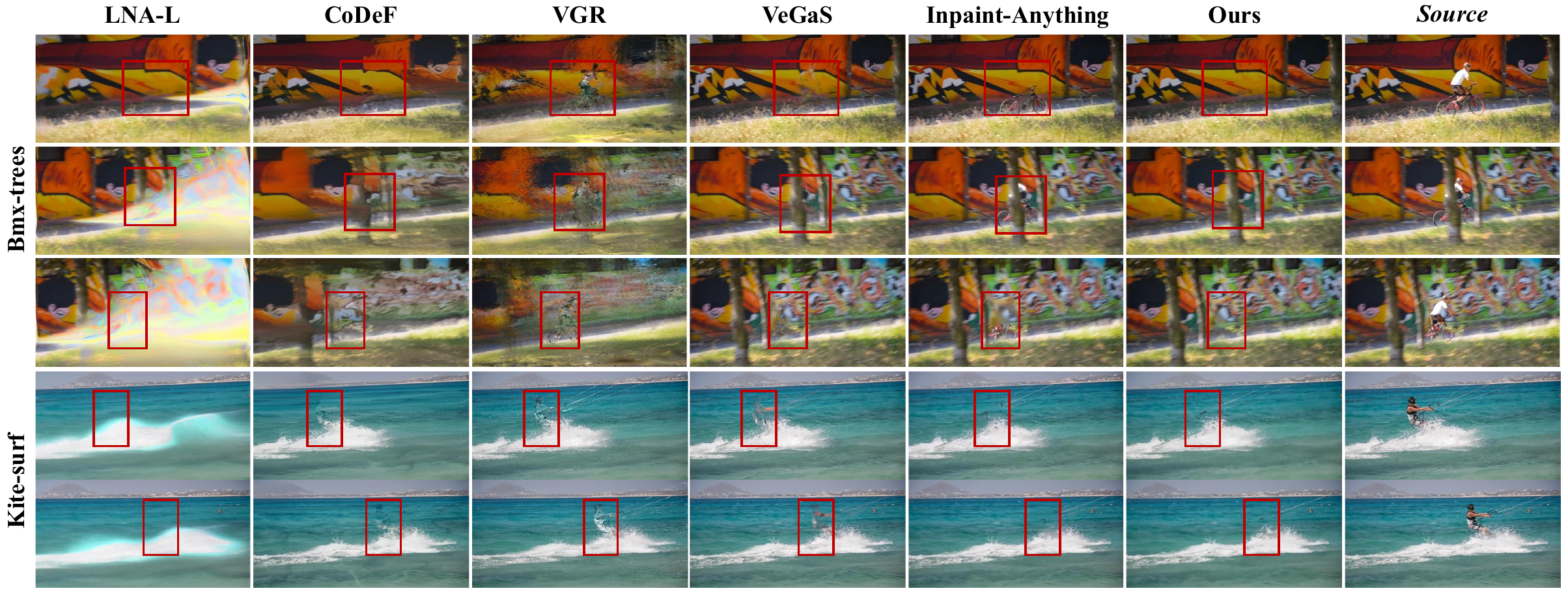} 
\vspace{-0.4cm}
\caption{Qualitative results of video in-painting. Our method effectively completes the background even during complex scene transitions. Please zoom in for detailed comparisons. More visualizations are provided in the supplementary material.}
\label{fig4:inpaint}
\vspace{-0.3cm}
\end{figure*}
\noindent\textbf{Video in-painting.} Thanks to our hierarchical vectorization strategy and the dynamic proxy propagation mechanism, we can achieve foreground removal and background completion by simply discarding the foreground proxy nodes during rendering. For comparison methods, LNA and CoDeF also adopt layered representations, which allow video in-painting by directly removing foreground layers. As for VeGaS and VGR, we modify their source codes to perform in-painting by masking out specific regions and removing the associated Gaussian primitives. Qualitative results are shown in Fig.~\ref{fig4:inpaint}, with more examples provided in the supplementary materials. As shown, even in the presence of large-scale motion and abrupt scene transitions, our method is capable of accurately removing the foreground and completing the background, producing smooth and temporally consistent in-painting results. In contrast, VGR and VeGaS exhibit noticeable artifacts due to the heavy stacking and coupling of Gaussian primitives. Although LNA and CoDeF adopt layered representations, they still struggle with large scene transitions due to their lack of intricate temporal modeling. For video inpainting algorithms based on generative models, it is evident that these methods tend to overfit the input video at the pixel-level representation. Consequently, they struggle to effectively capture the underlying appearance and motion dynamics. As a result, the inpainted regions often exhibit noticeable structural inconsistencies and visual artifacts.

\begin{figure*}[t]
\centering
\includegraphics[width=1.0\linewidth]{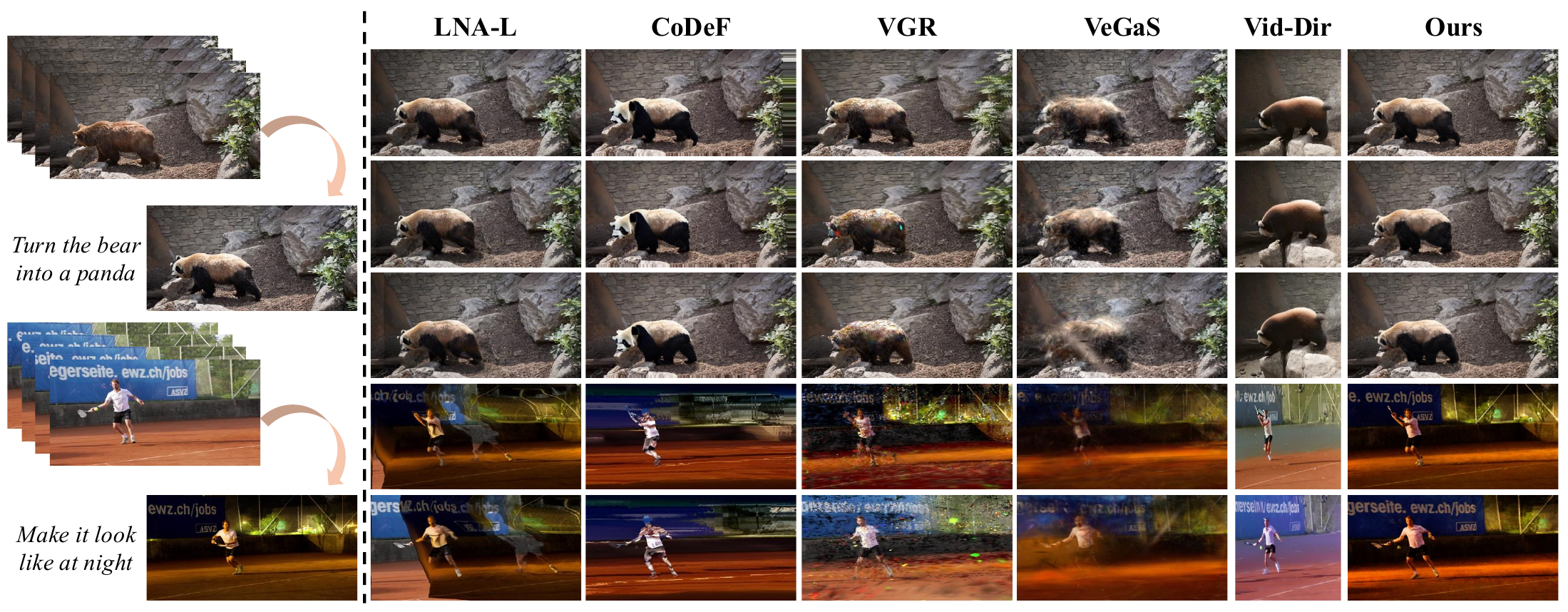} 
\caption{Qualitative results of image-based consistent editing. Our method ensures stable and controlled image editing propagation across the entire video, even in cases of large-scale motion, outperforming compared methods. Please zoom in for details.}
\label{fig5:editing}
\end{figure*}

\noindent\textbf{Image-based Consistent Editing.}
Since our representation decouples all information in the video and stores it in a distributed manner across proxy nodes, we can apply image editing algorithms (we use InstructP2P~\cite{brooks2023instructpix2pix} in this work) to video editing tasks by re-optimizing the features on the proxy nodes corresponding to the region of editing interest. VGR and VeGaS also perform video editing by re-optimizing Gaussian parameters on the edited images. LNA and CoDeF propagate the editing information across the entire video by editing atlases and the canonical image. However, as shown in Fig.~\ref{fig5:editing}, LNA and CoDeF struggle to handle large non-rigid motions, as they lead to severe distortions of atlases and the canonical image. Similarly, both VGR and VeGaS fail to produce satisfactory results due to the excessive accumulation of Gaussian primitives and the lack of precise temporal correlations between the primitives.
In contrast, our method benefits from the stable and temporally consistent representation of local structure and texture information via proxy nodes, enabling the stable and accurate propagation of image edits throughout the video. We also compare our method with an advanced generative model-based video editing approach Vid-Dir~\cite{wang2025videodirector}. As shown, current video generation models struggle to handle out-of-distribution data effectively and often produce unexpected results, such as unintended edits in unrelated regions.

\begin{figure*}[t]
\centering
\includegraphics[width=0.95\linewidth]{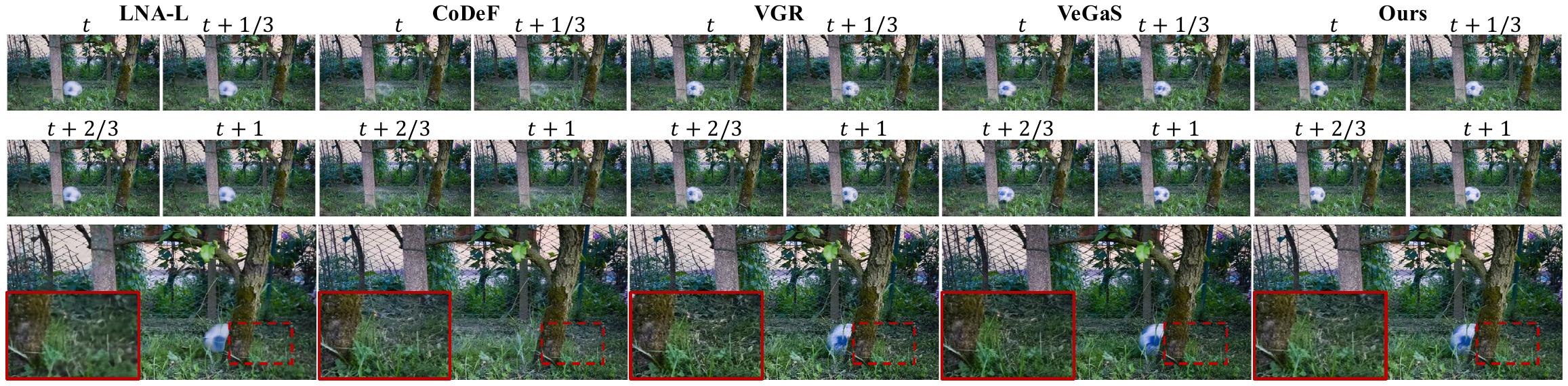} 
\caption{Qualitative results of Spatio-Temporal Interpolation. Our method preserves fine texture details during both spatial and temporal frame interpolation. Please zoom in for details.}
\label{fig6:interp}
\end{figure*}

\noindent\textbf{Spatio-Temporal Interpolation.}
Since our proxy nodes are distributed in continuous space, we can perform spatial video interpolation by simply increasing the number of pixels during rendering. Comparisons with other representations on the spatial interpolation task are presented in Fig.~\ref{fig6:interp} (bottom half). Note that our method preserves fine texture details even when trained at a lower resolution and rendered at a higher resolution thanks to the stable and continuous representation of local structures and textures provided by the proxy nodes. For the temporal interpolation task, we can freely adjust the video playback speed by simply performing continuous interpolation over time steps and remapping the trajectories of proxy nodes to the new temporal positions. The qualitative results are also shown in Fig.~\ref{fig6:interp} (upper half). We can see that our method enables smoother frame interpolation, producing high-quality intermediate frames without introducing flickering or motion artifacts.

\subsection{Ablation Study}
We conduct ablation studies on key components and hyperparameters of our framework. We report quantitative results on $480$p version DAVIS benchmark. Additional qualitative results are provided in supplementary materials.

\begin{table}[]
\small
    \centering
    \begin{tabular}{l|c|c|c|c|c}
    \toprule
    ~& w/o-layer & F & F\&L & w/o-pos & w/o-$\mathcal{U}$ \\
    \midrule
    PSNR &\textbf{30.78}&28.51&29.27&29.52&27.03  \\
    Params.&3.12M&3.11M&3.14M&2.99M&\textbf{0.65M} \\
    \bottomrule
    \end{tabular}
    \caption{\textbf{Component Analyses.} ``w/o-layer'' removes spatial vectorization, processing the video as a whole without semantic decomposition. ``F'' and ``F\&L'' sample proxy nodes only from the first frame, or from the first and last frames, respectively. ``w/o-pos'' and ``w/o-$\small \mathcal{U}$'' disable position input and high-frequency embedding in the implicit texture representation. }
    \label{tab:ablation}
\end{table}

\begin{figure}
\vspace{-0.2cm}
\includegraphics[width=0.95\linewidth]{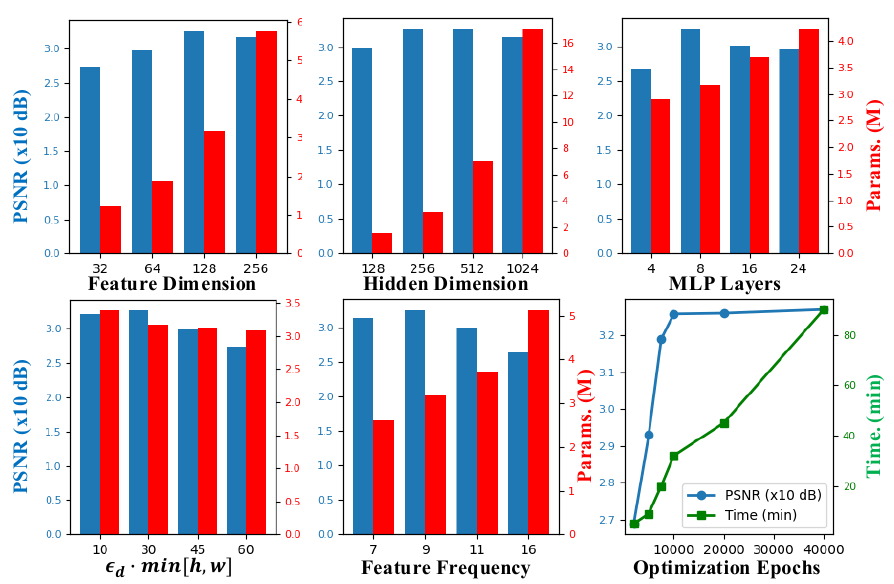} 
\vspace{-0.2cm}
\caption{Parameter Analyses.}
\label{fig7:ablate}
\vspace{-0.4cm}
\end{figure}

\noindent\textbf{Component Analyses.}  As shown in Tab.~\ref{tab:ablation}, our method achieves competitive results even without semantic layering, demonstrating its robustness. ``F'' and ``F\&L'' perform poorly on long sequences due to their limitations in handling large motions and scene changes. In contrast, our dynamic proxy update mechanism addresses these challenges with minimal parameter overhead. Additionally, while high-frequency encoding of features and coordinates significantly increases parameter count, it indeed greatly enhances the model’s ability to capture video appearance.

\noindent\textbf{Parameter Analyses.} We investigate the impact of MLP architecture and hyperparameter ($\epsilon_d$ and training epochs) settings on reconstruction performance, as shown in Fig.~\ref{fig7:ablate}. Note that we follow the principle of balancing representational quality and efficiency when determining all the parameters in our experiments.

\section{Conclusion}
This work introduces a novel and efficient video representation that simultaneously embeds motion and appearance into hierarchical spatio-temporally consistent proxy nodes. Extensive experiments on various tasks demonstrate that our representation effectively reconstructs complex videos with significant parameter compression and supports complex video processing tasks, even in highly complex video scenarios with large-scale motion and frequent scene changes.


{
    \small
    \bibliographystyle{ieeenat_fullname}
    \bibliography{main}
}

\clearpage
\setcounter{page}{1}
\maketitlesupplementary
\section{Supplementary Materials}
In the supplementary materials, we provide more detailed experimental setups and more detailed qualitative results on video reconstruction, video inpainting and image-based consistent editing tasks in this pdf.

\subsection{Experimental setups}
\noindent\textbf{Benchmark.} We mention in the paper that we use all videos from both resolution variants of the DAVIS dataset as our benchmark. However, for LNA, we only sample a subset of DAVIS for test due to the extremely long optimization time (over $24$hours for one video). The sampled list is shown as below:
\begin{table}[H]
\small
    \centering
    \begin{tabular}{l|l|l|l}
    \toprule
    bear & blackswan & bmx-trees & boat  \\
    breakdance-flare & bmx-bumps & car-turn & dog-agility  \\
    drift-straight & flamingo & giraffe  & kite-surf \\
    libby & drift-chicane & motorbike & paragliding  \\
    breakdance & lucia & horsejump-low & parkour \\
    scooter-black & soccerball & swing & tennis \\
    \bottomrule
    \end{tabular}
    \caption{Sampled videos from DAVIS for computing metrics for LNA due to its extremelt long optimization time.}
    \label{tab:supp_LNA}
\end{table}

\begin{figure}[H]
    \centering
    \includegraphics[width=1.0\linewidth]{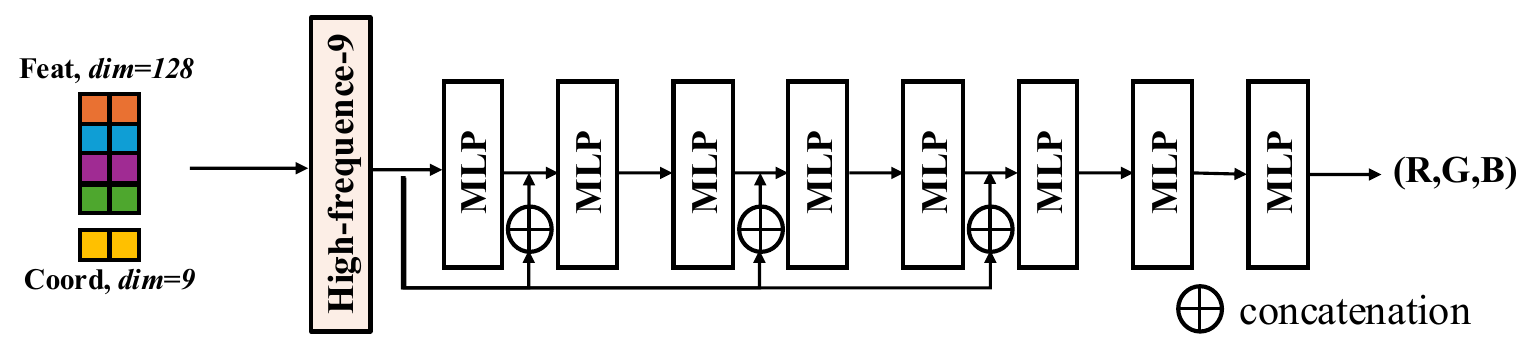}
    \caption{MLP architecture of our implicit appearance representation.}
    \label{fig:supp_mlp}
\end{figure}

\noindent\textbf{Our architecture.} We show the pipeline of our method in Fig.~\ref{fig2:overview} of the paper. Here in the supplementary material, we provide the detailed architecture of our network for implicit appearance modeling, as shown in Fig.~\ref{fig:supp_mlp}.

\subsection{More Results}
\subsubsection{Video Representation}
\begin{figure*}[t]
\centering
\includegraphics[width=1.0\linewidth]{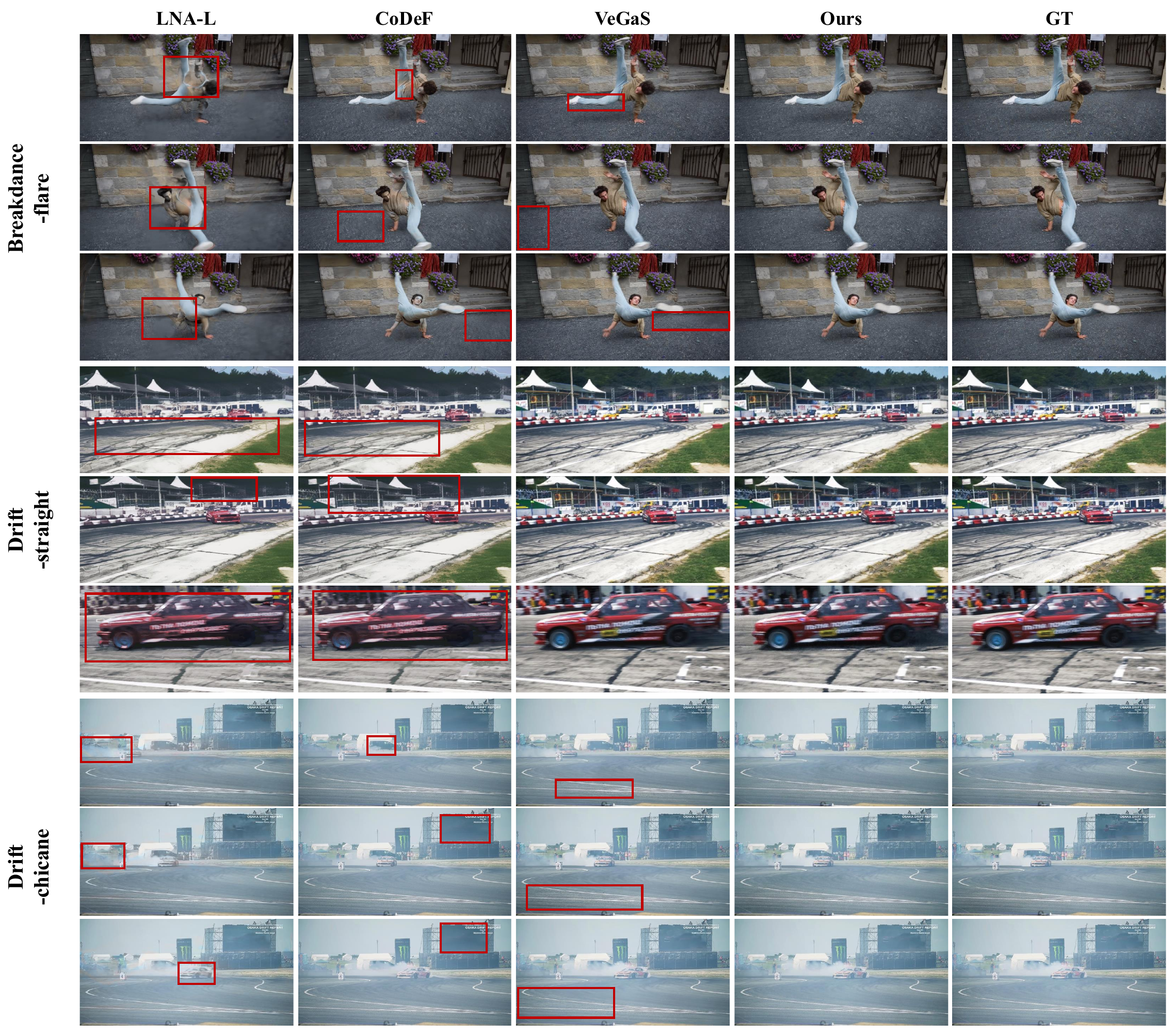} 
\caption{Qualitative comparisons with sota methods on \textbf{Video Reconstruction} task. We highlight the critical distortions of the comparison methods with red bounding boxes. Please zoom in for detailed comparisons.}
\label{fig:supp_recons}
\end{figure*}

We observe that, as shown in Tab.~\ref{tab:video_recons}, only VeGaS and our proposed method achieve satisfactory performance on the DAVIS dataset. To further evaluate their generalization capability, we conduct an additional comparison between our method and VeGaS on a new dataset FBMS~\cite{ochs2013segmentation}, with the results summarized in Tab.~\ref{tab:video_recons2}. More qualitative reconstruction results are shown in Fig.~\ref{fig:supp_recons}.
\begin{table}
\small
    \centering
    \begin{tabular}{l|c|c|c|c|c}
    \toprule
    Method &PSNR$\uparrow$&LPIPS$\downarrow$&SSIM$\uparrow$&Params.$\downarrow$&Time$\downarrow$ \\
    \midrule
    VeGaS~\cite{smolak2024vegas} & 32.82 & 0.1203 & 0.9003 & 25.62M & 1h \\
    Ours & \textbf{33.56} & \textbf{0.0852} & \textbf{0.9092} & \textbf{3.17M} & 30min \\
    \bottomrule
    \end{tabular}
    \caption{\textbf{Results on FBMS~\cite{ochs2013segmentation} dataset.} We make comparisons with state-of-the-art video representation VeGaS. The quantitative results demonstrate that our representation achieves superior video reconstruction quality while utilizing fewer parameters.}
    \label{tab:video_recons2}
\end{table}

\subsubsection{Video Inpainting}
\begin{figure*}[t]
\centering
\includegraphics[width=1.0\linewidth]{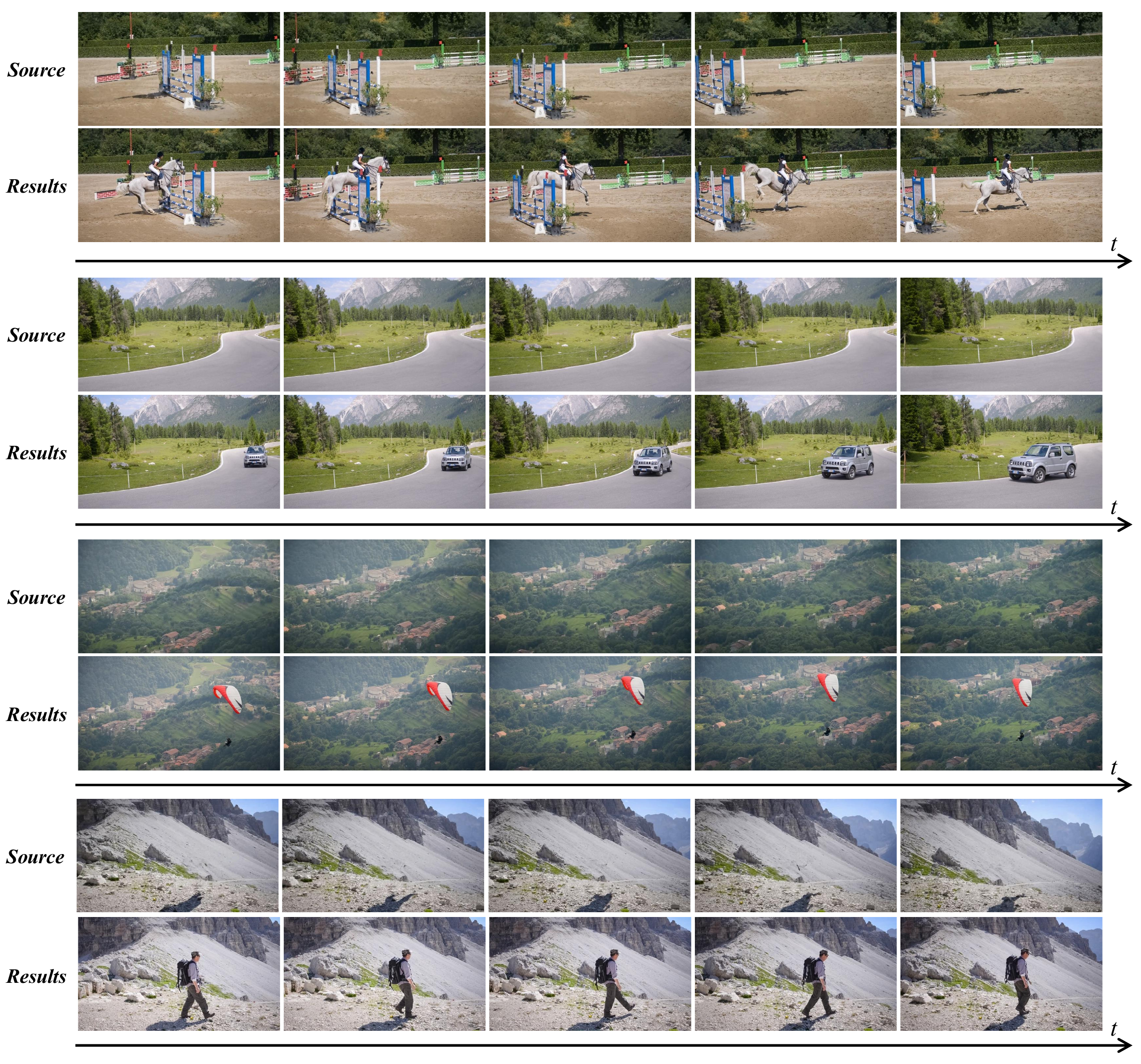} 
\caption{More Qualitative results on \textbf{Video Inpainting} task.}
\label{fig:inpaint_supp}
\end{figure*}

We provide additional visual examples of video inpainting. As illustrated in Fig.~\ref{fig:inpaint_supp}, our proxy representation is capable of completing regions occluded by the foreground even in cases involving large motions or significant scene changes. Notably, this is achieved solely by leveraging video priors, without relying on any generative model.

\subsubsection{Video Editing}
\begin{figure*}[t]
\centering
\includegraphics[width=1.0\linewidth]{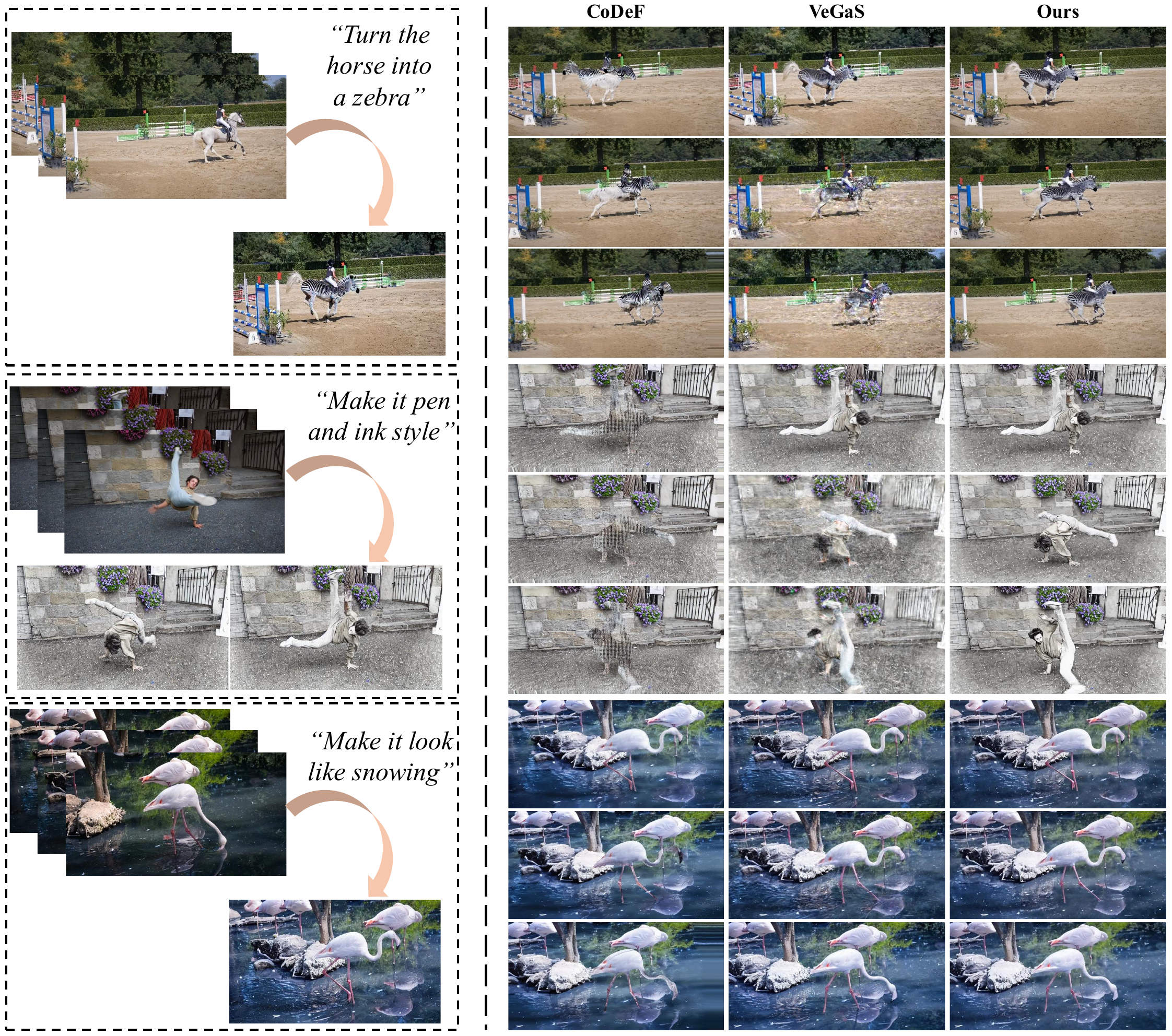} 
\caption{Qualitative comparisons with sota methods on \textbf{Video Editing} task. Please zoom in for detailed comparisons.}
\label{fig:supp_edit}
\end{figure*}

\begin{figure*}[t]
\centering
\includegraphics[width=1.0\linewidth]{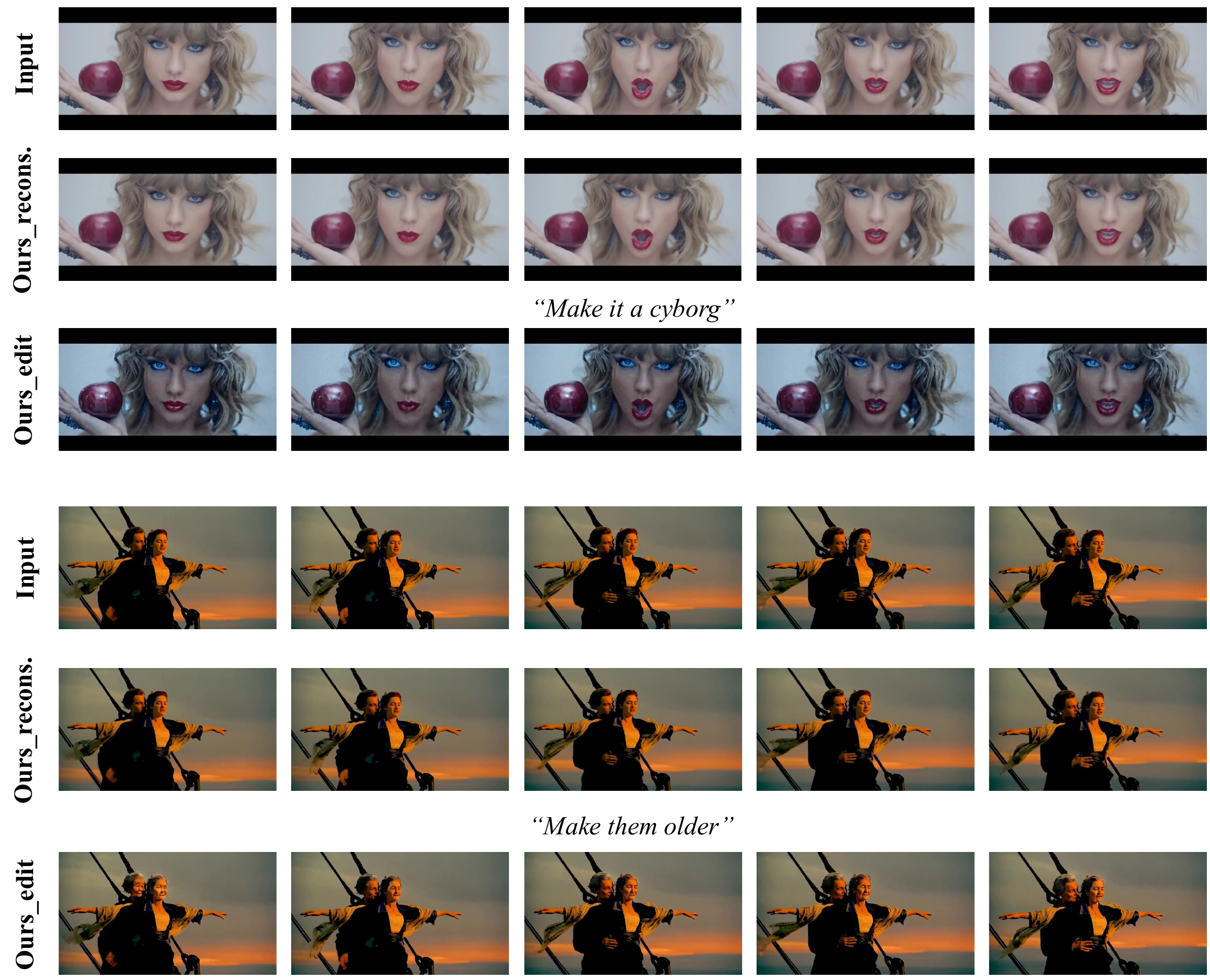} 
\caption{Qualitative results of our method on In-the-wild Videos. Please zoom in for more details.}
\label{fig:supp_newrecons}
\end{figure*}

In Fig.~\ref{fig:supp_edit} and Fig.~\ref{fig:supp_newrecons}, we present additional video-editing results. As shown, our representation can stably and accurately propagate edits made on a single image to the entire video. 

\end{document}